\documentclass{article}
\usepackage{ijcai25}

\usepackage{times}
\usepackage{soul}
\usepackage{url}
\usepackage[hidelinks]{hyperref}
\usepackage[utf8]{inputenc}
\usepackage[small]{caption}
\usepackage{graphicx}
\usepackage{amsmath}
\usepackage{amsthm}
\usepackage{booktabs}
\usepackage{algorithm}
\usepackage{algorithmic}
\usepackage[switch]{lineno}
\usepackage{amssymb}
\usepackage{amsfonts}
\usepackage{gensymb}
\usepackage{multirow}
\usepackage[table,xcdraw]{xcolor}
\usepackage{wasysym}
\usepackage{booktabs}

\title{BEVTrack: A Simple and Strong Baseline for 3D Single Object \\Tracking in Bird's-Eye View}

\author{
Yuxiang Yang$^1$,
Yingqi Deng$^1$,
Mian Pan$^1$,
Zheng-Jun Zha$^2$$^*$,
Jing Zhang$^3$\thanks{Corresponding author.}\\
\affiliations
$^1$School of Electronics and Information, Hangzhou Dianzi University, China\\
$^2$Department of Automation, University of Science and Technology of China, China\\
$^3$School of Computer Science, Wuhan University, China\\
\emails
\{yyx, den, ai\}@hdu.edu.cn, zhazj@ustc.edu.cn, jingzhang.cv@gmail.com
}

\begin{document}

\maketitle

\begin{abstract}

3D Single Object Tracking (SOT) is a fundamental task in computer vision and plays a critical role in applications like autonomous driving. However, existing algorithms often involve complex designs and multiple loss functions, making model training and deployment challenging. Furthermore, their reliance on fixed probability distribution assumptions (e.g., Laplacian or Gaussian) hinders their ability to adapt to diverse target characteristics such as varying sizes and motion patterns, ultimately affecting tracking precision and robustness. To address these issues, we propose BEVTrack, a simple yet effective motion-based tracking method. BEVTrack directly estimates object motion in Bird's-Eye View (BEV) using a single regression loss. To enhance accuracy for targets with diverse attributes, it learns adaptive likelihood functions tailored to individual targets, avoiding the limitations of fixed distribution assumptions in previous methods. This approach provides valuable priors for tracking and significantly boosts performance. Comprehensive experiments on KITTI, NuScenes, and Waymo Open Dataset demonstrate that BEVTrack achieves state-of-the-art results while operating at 200 FPS, enabling real-time applicability. The code will be released at \href{https://github.com/xmm-prio/BEVTrack}{https://github.com/xmm-prio/BEVTrack}.

\end{abstract}

\section{Introduction}
\label{sec:intro}

\begin{figure}[tp]
    \centering
    \includegraphics[width=\linewidth]{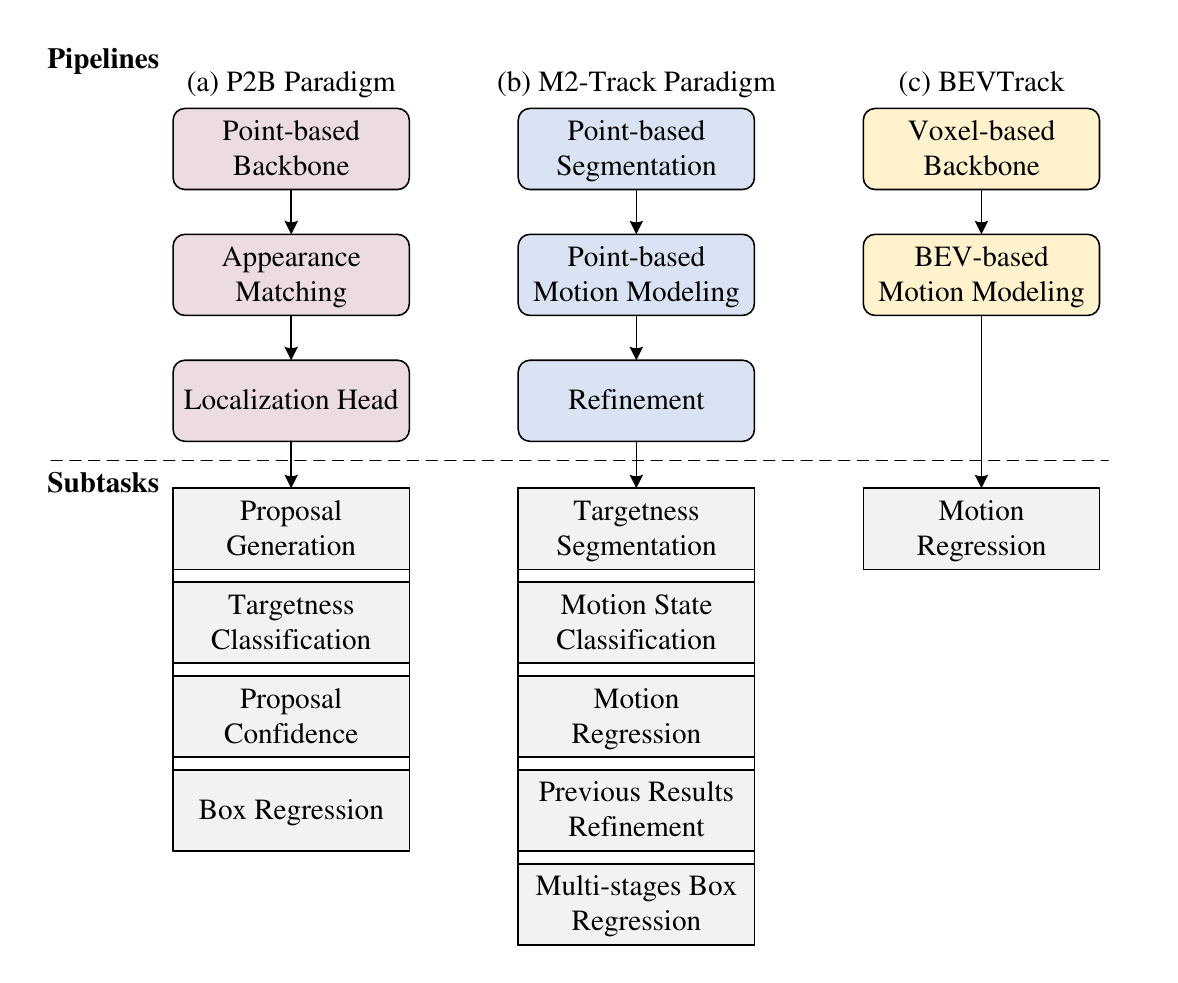}
    \caption{Comparison with typical 3D SOT paradigms. Previous methods mainly rely on point-based representations, and decompose the tracking problem into multiple subtasks, leading to a complicated tracking framework. On the contrary, our proposed BEVTrack simplifies the tracking pipeline with a single regression loss.} 
    \label{fig:paradigm}
\end{figure}

3D single object tracking (SOT) is crucial for various applications, such as autonomous driving~\cite{centerpoint,crossd}. It aims to localize a specific target across a sequence of point clouds, given only its initial status. Existing tracking approaches~\cite{P2B,M2Track} commonly adopt point-based representations, directly taking raw point clouds as input. For example, P2B~\cite{P2B} and its follow-up works~\cite{BAT,PTTR} adopt a point-based network~\cite{PointNet++,DGCNN} with the Siamese architecture for feature extraction, followed by a point-based appearance matching module~\cite{P2B,PTTR,CXTrack} for propagation of target cues, and a 3D Region Proposal Network~\cite{VoteNet,3dSiamRPN} for target localization, as illustrated in Fig.~\ref{fig:paradigm}(a). M2-Track~\cite{M2Track} proposes a motion-centric paradigm, that first segments the target points from their surroundings with a PointNet~\cite{PointNet} segmentation network and then localizes the target through a motion modeling approach followed by a box refinement module, as illustrated in Fig.~\ref{fig:paradigm}(b).

%%picture%%

Although these approaches have demonstrated strong performance on tracking benchmarks, their reliance on intricate designs and the need to address multiple subtasks significantly increase framework complexity, making both training and deployment challenging. This raises an important question: can we simplify the tracking formulation while maintaining competitive performance?

In this paper, we present BEVTrack, a simple yet strong baseline for 3D SOT, as shown in Fig.~\ref{fig:paradigm}(c). By estimating the target motion in Bird's-Eye View (BEV) to perform tracking, BEVTrack demonstrates surprising simplicity from various aspects, \textit{i.e.}, network designs, training objectives, and tracking pipeline. Specifically, we first adopt a voxel-based network~\cite{VoxelNet} with the Siamese architecture for feature extraction. Subsequently, we compress height information into the channel dimension to obtain the BEV features. Given that corresponding objects are spatially adjacent in the BEV features across consecutive frames, we can easily fuse them together with an element-wise operation such as concatenation~\cite{pillarflownet}. Then we adopt several convolutional and down-sampling layers to capture the object motion with a wide range of patterns. Finally, we discard the complicated region proposal network while using a global max pooling followed by a lightweight multilayer perception (MLP) to regress the relative target motion.

To optimize the regression-based trackers, current approaches typically employ conventional $L_{1}$ or $L_{2}$ loss. This kind of supervision actually makes a fixed Laplacian or Gaussian assumption on the data distribution, which is inflexible when handling targets possessing diverse attributes (\textit{e.g.}, sizes, moving patterns, and sparsity degrees). For example, when tracking targets with high sparsity degrees, where the predictions may be uncertain while the annotations are not very reliable, the outputs should conform to a distribution with a large variance. To this end, we introduce a novel distribution-aware regression strategy for tracking, which constructs the likelihood function with the learned underlying distributions adapted to distinct targets, instead of making a fixed assumption. Note that this strategy does not participate in the inference phase, thus boosting the tracking performance without additional computation overhead. Despite no elaborate design, BEVTrack exhibits a substantial performance advantage over the current state-of-the-art (SOTA) methods on challenging tracking datasets, \textit{i.e.}, KITTI~\cite{KITTI}, NuScenes~\cite{NuScenes}, and Waymo Open Dataset~\cite{waymo}, while operating at 200 FPS, enabling real-time applicability.

The main contributions of this paper are three-fold:

$\bullet$ We propose BEVTrack, a simple yet strong baseline for 3D SOT. BEVTrack marks the first to perform tracking through motion modeling in BEV, resulting in a simplified tracking pipeline design.

$\bullet$ We present a novel distribution-aware regression strategy for tracking, which constructs the likelihood function with the learned underlying distributions adapted to targets possessing diverse attributes. This strategy provides accurate guidance for tracking, resulting in improved performance while avoiding extra computation overhead.

$\bullet$ BEVTrack achieves SOTA performance on three popular benchmarks while maintaining a high inference speed.

\section{Related Work}
%\subsection{3D Single Object Tracking} 
Early 3D SOT approaches~\cite{rgbd2,rgbd3} predominantly utilize RGB-D data and often adapt 2D Siamese networks by integrating depth maps. However, variations in illumination and appearance can adversely affect the performance of such RGB-D techniques. SC3D~\cite{sc3d} is the first 3D Siamese tracker based on shape completion that generates a large number of candidates in the search area and compares them with the cropped template, taking the most similar candidate as the tracking result. The pipeline relies on heuristic sampling and does not learn end-to-end, which is very time-consuming. 

To address these issues, P2B~\cite{P2B} develops an end-to-end framework by employing a shared point-based backbone for feature extraction, followed by a point-wise appearance-matching technique for target cues propagation. Ultimately, a Region Proposal Network is used to derive 3D proposals, of which the one with the peak score is selected as the final output. P2B reaches a balance between performance and efficiency, and many works~\cite{BAT,PTT,PTTR,STNet} follow the same paradigm. Drawing inspiration from the success of transformers~\cite{transformer}, many studies have incorporated elaborate attention mechanisms to enhance feature extraction and target-specific propagation. For instance, PTT~\cite{PTT} introduces the Point Track Transformer to enhance point features. PTTR~\cite{PTTR} presents the Point Relation Transformer for target-specific propagation and a Prediction Refinement Module for precision localization. Similarly, ST-Net~\cite{STNet} puts forth an iterative correlation network, and CXTrack~\cite{CXTrack} presents the Target Centric Transformer to harness contextual information from consecutive frames. SyncTrack~\cite{synctrack} introduces a single-branch and single-stage framework, without Siamese-like forward propagation and a standalone matching network. MBPTrack~\cite{mbptrack} exploits both spatial and temporal contextual information using a memory mechanism. Unlike the Siamese paradigm, M2-Track~\cite{M2Track} proposes a motion-centric paradigm, explicitly modeling the target's motion between two successive frames. This motion-centric paradigm shows robustness to the problems caused by the appearance variation and the sparsity of point clouds. Our study also adopts a motion-centric paradigm but focuses on simplifying the tracking formulation, resulting in a streamlined and efficient solution that maintains strong performance.

\section{Method}

\begin{figure*}[tp]
    \includegraphics[width=0.9\textwidth]{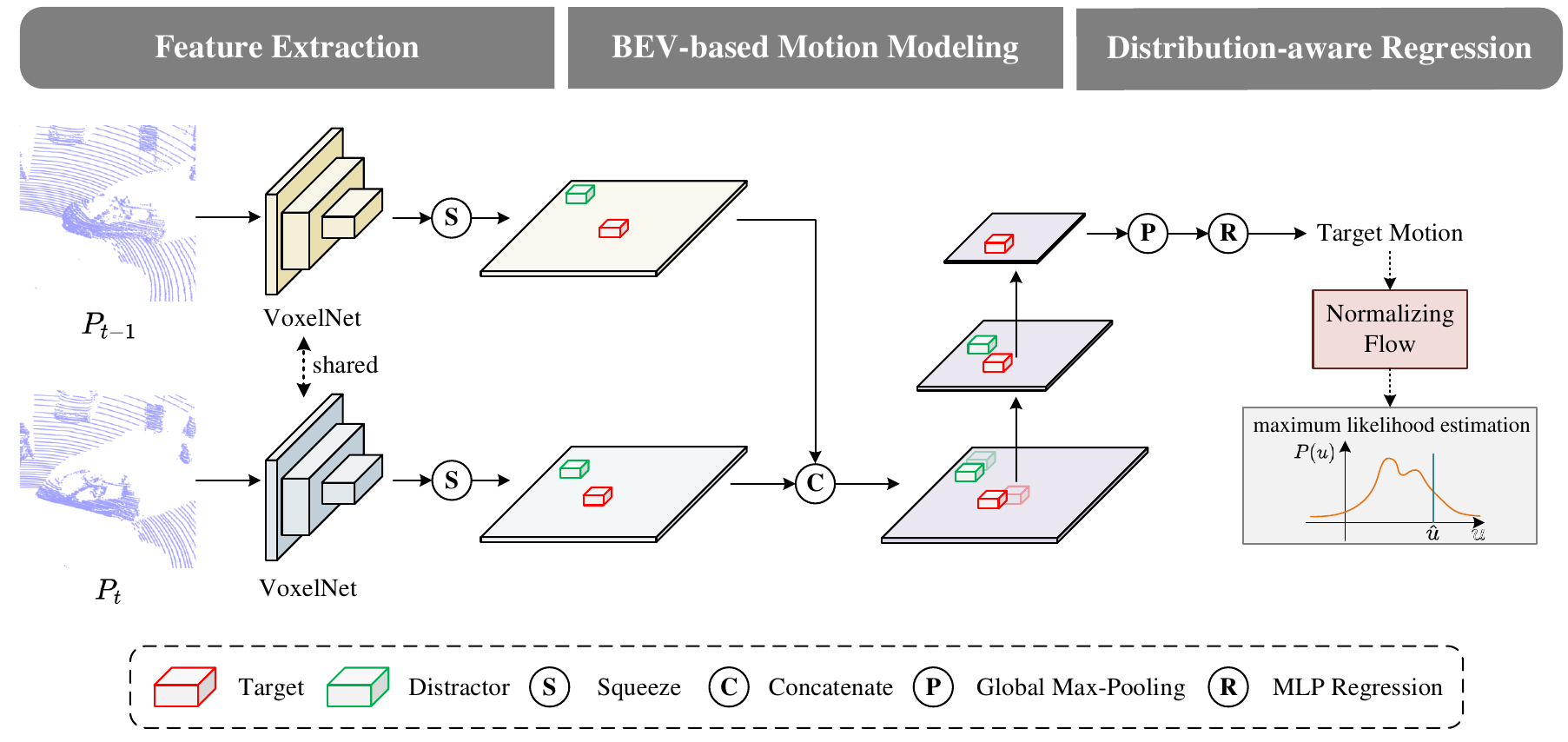}
    \centering
    \caption{Architecture of BEVTrack. The proposed framework contains three parts including voxel-based feature extraction, BEV-based motion modeling, and distribution-aware regression. Our BEVTrack is a simple tracking baseline framework with a plain convolutional architecture and a single regression loss, yet demonstrating state-of-the-art performance. }
    \label{fig:bevtrack}
\end{figure*}

\subsection{Overview}
Given the 3D bounding box (BBox) of a specific target at the initial frame, 3D SOT aims to localize the target by predicting its 3D BBoxes in the subsequent frames. A 3D BBox $\mathcal{B}_{t} \in \mathbb{R}^{7}$ is parameterized by its center (\textit{i.e.}, $x,y,z$ coordinates), orientation (\textit{i.e.}, heading angle $\theta$ around the $up$-axis), and size (\textit{i.e.}, width, length, and height). Suppose the point clouds in two consecutive frames are denoted as $\mathcal{P}_{t-1} \in \mathbb{R}^{N_{t-1}\times 3}$ and $\mathcal{P}_{t} \in \mathbb{R}^{N_{t}\times 3}$, respectively, where $N_{t-1}$ and $N_{t}$ are the numbers of points in the point clouds. Notably, the two point clouds have been transformed to the canonical system w.r.t the target BBox $\mathcal{B}_{t-1}$ (either given as the initial state or predicted by the tracker). Since the size of the tracking target rarely changes, we follow the common practice~\cite{P2B} and assume a constant target size. Consequently, the inter-frame target translation offsets (\textit{i.e.}, $\Delta x, \Delta y, \Delta z$) and the yaw angle offset $\Delta\theta$ are regressed. By applying the translation and rotation to the 3D BBox $\mathcal{B}_{t-1}$, we can compute the 3D BBox $\mathcal{B}_{t}$ to localize the target in the current frame. The tracking process can be formulated as:
\begin{equation}
  \mathcal{F(P_{\mathit{t-1}}, P_{\mathit{t}})} \mapsto (\Delta x, \Delta y, \Delta z, \Delta\theta),
  \label{eq:definition}
\end{equation}
where $\mathcal{F}$ is the mapping function learned by the tracker.

Following Eq.~\eqref{eq:definition}, we propose BEVTrack, a simple yet strong baseline for 3D SOT. The overall architecture of BEVTrack is presented in Fig.~\ref{fig:bevtrack}. It first employs a shared voxel-based backbone~\cite{VoxelNet} to extract 3D features, which are then squeezed to derive the BEV features (Sec.~\ref{subsec:featureextraction}). Subsequently, BEVTrack fuses the BEV features via a concatenation operation and several convolutional layers (Sec.~\ref{subsec:bmm}). Finally, it adopts a simple prediction head which contains a global max pooling layer followed by a lightweight MLP. For prediction, we employ a novel distribution-aware regression strategy in the training phase (Sec.~\ref{sec:loss}). 

\subsection{Feature Extraction}
\label{subsec:featureextraction}

To localize the target from surroundings accurately, we need to learn discriminative features from the point clouds. Instead of using a point-based backbone~\cite{PointNet,PointNet++,DGCNN} as in~\cite{P2B,M2Track,CXTrack}, we adopt VoxelNet~\cite{VoxelNet} as the shared backbone. Specifically, we voxelize the 3D space into regular voxels and extract the voxel features of each non-empty voxel by a stack of sparse convolutions, where the initial feature of each voxel is simply calculated as the mean values of point coordinates within each voxel in the canonical coordinate system. The spatial resolution is downsampled 8 times by three sparse convolutions~\cite{sparse_conv} of stride 2, each of which is followed by several submanifold sparse convolutions~\cite{submanifold}. Afterwards, we squeeze the sparse 3D features along the height dimension to derive the BEV features $\mathcal{X}_{t-1} \in \mathbb{R}^{H \times W \times C}$ and $\mathcal{X}_{t} \in \mathbb{R}^{H \times W \times C}$, where $H$ and $W$ denote the 2D grid dimension and $C$ is the number of feature channels.

\begin{figure}[tp]
    \includegraphics[width=\linewidth]{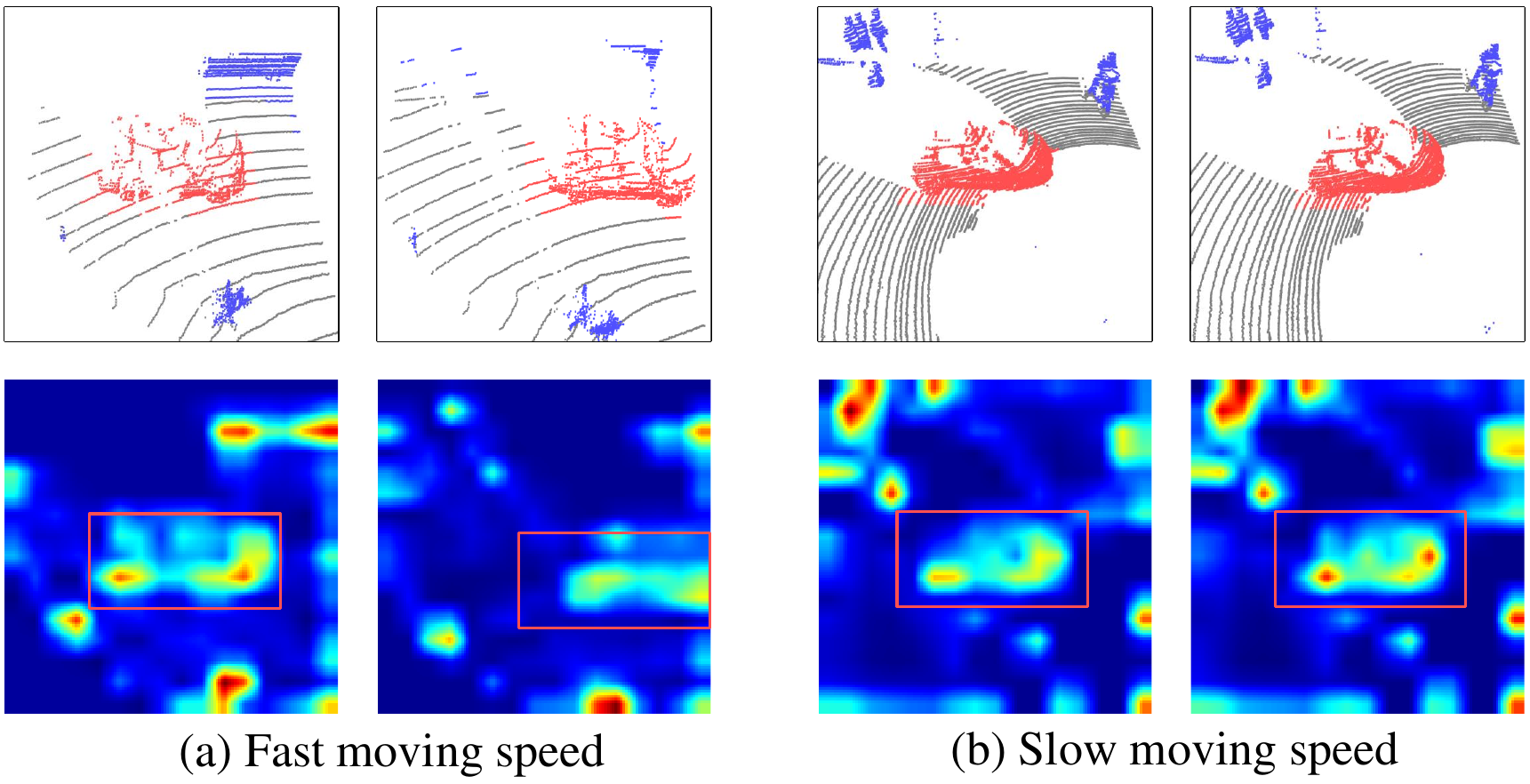}
    \centering
    \caption{The motion pattern of objects is variable in different scenes. The top rows show point cloud scenes across two consecutive frames, where the \textcolor{red}{red} points indicate the target. The bottom rows visualize the heatmaps of the BEV response map with ground-truth bounding box (in \textcolor{red}{red} rectangles).}
    \label{fig:motion}
\end{figure}

\subsection{BEV-based Motion Modeling}
\label{subsec:bmm}

BEV-based Motion Modeling (BMM) aims to model the motion relations of the target from consecutive frames. Given the BEV features $\mathcal{X}_{t-1}$ and $\mathcal{X}_{t}$, the corresponding objects are spatially adjacent across them. Therefore, we can easily fuse them together with an element-wise operator such as concatenation to preserve their spatial proximity. Furthermore, we apply several convolutional blocks to encode their motion relations. It is noteworthy that the receptive field of the convolution is limited due to the small kernel size, while the motion pattern of objects is variable in different scenes. As shown in Fig.~\ref{fig:motion}, objects with a fast-moving speed are far apart across the BEV feature maps. To deal with this issue, we propose to enlarge the receptive field through spatial down-sampling by inserting convolutions of stride 2 at intervals, allowing it to capture a wide range of motion patterns. The above process can be formulated as:
\begin{equation}
  \mathcal{Y}=\text{Convs}([\mathcal{X}_{t-1};\mathcal{X}_{t}]),
\end{equation}
where $\text{Convs}$ denotes the convolutional blocks in BMM and $[;]$ denotes the concatenation operator. $\mathcal{Y} \in \mathbb{R}^{H' \times W' \times C'}$, where $H'$, $W'$, and $C'$ denote the spatial dimension and the number of feature channels, respectively. See the appendix for the ablation experiments on the design choices of BMM.

\subsection{Distribution-aware Regression}
\label{sec:loss}
Different from the existing two-stage point-to-box prediction head~\cite{P2B,PTTR,STNet}, which contains two parts of proposals generation and proposal-wise scores prediction, we propose a one-stage post-processing-free prediction head consisting of only a global max-pooling layer and an MLP, \textit{i.e.}, 
\begin{equation}
  \mathcal{M}=\text{MLP}(\text{Pool}(\mathcal{Y})),
\end{equation}
where $\mathcal{M} \in \mathbb{R}^{8}$ denotes the expectation of the target translation and yaw offsets $\bar{u} \in \mathbb{R}^{4}$ and their standard deviation $\sigma \in \mathbb{R}^{4}$, in contrast to prior methods that solely regress the deterministic target motion $u$. By applying the predicted translation to the last state of the target, we can localize the target in the current frame. Note that the standard deviation $\sigma$ enables dynamic optimization during training while being removed in the inference phase, as elaborated below.

The difference in sizes, moving patterns, and sparsity degrees among the tracked targets poses great challenges to existing trackers. To address this issue, we propose a novel distribution-aware regression strategy for tracking, which constructs the likelihood function with the learned underlying distributions adapted to distinct targets. In this way, the model can adaptively handle targets with different attributes, thus improving tracking performance.

Following~\cite{RLE}, we model the distribution of the target motion $u \sim P(u)$ with reparameterization, which assumes that objects belonging to the same category share the same density function family, but with different mean and variance. Specifically, $P(u)$ can be obtained by scaling and shifting $z$ from a zero-mean distribution $z \sim P_{Z}(z)$ with transformation function $u=\bar{u}+\sigma \cdot z$, where $\bar{u}$ denotes the expectation of the target translation offsets and $\sigma$ denotes the scale factor of the distribution. $P_{Z}(z)$ can be modeled by a normalizing flow model (\textit{e.g.}, real NVP~\cite{NVP}). Given this transformation function, the density function of $P(u)$ can be calculated as:
\begin{equation}
  \log P(u)=\log P_{Z}(z)-\log \sigma.
  \label{eq:density}
\end{equation}

In this work, we employ residual log-likelihood estimation~\cite{RLE} to estimate the above parameters, which factorizes the distribution $P_{Z}(z)$ into one prior distribution $Q_{Z}(z)$ (\textit{e.g.}, Laplacian distribution or Gaussian distribution) and one learned distribution $G_{Z}(z \mid \theta)$. To maximize the likelihood in Eq.~\eqref{eq:density}, we can minimize the following loss:
\begin{equation}
  \mathcal{L} =-\log Q_{Z}(\hat{z})-\log G_{Z}(\hat{z} \mid \theta)+\log \sigma.
  \label{eq:rle}
\end{equation}
Here, $\hat{z}=(\hat{u}-\bar{u})/ \sigma$, $\hat{u}$ is the ground truth translation offsets.

\section{Experiment}
\subsection{Experimental Settings}

We validate the effectiveness of BEVTrack on three widely-used challenging datasets: KITTI~\cite{KITTI}, NuScenes~\cite{NuScenes}, and Waymo Open Dataset (WOD)~\cite{waymo}. KITTI contains 21 video sequences for training and 29 video sequences for testing. We follow previous work~\cite{P2B} to split the training set into train/val/test splits due to the inaccessibility of the labels of the test set. NuScenes contains 1,000 scenes, which are divided into 700/150/150 scenes for train/val/test. Following the implementation in~\cite{BAT}, we compare with the previous methods on five categories including \textit{Car}, \textit{Pedestrian}, \textit{Truck}, \textit{Trailer}, and \textit{Bus}. WOD includes 1150 scenes with 798 for training, 202 for validation, and 150 for testing. Following M2-Track~\cite{M2Track}, We conduct training and testing respectively on the training and validation set. Following~\cite{P2B}, we adopt Success and Precision defined in one pass evaluation (OPE)~\cite{metrics} as the evaluation metrics.

\begin{table}[!t]
%\captionsetup{justification=centering}
\centering
\resizebox{\linewidth}{!}{
\renewcommand{\arraystretch}{1.1}
\begin{tabular}{c|c|c|c|c|l}
\hline
Method &
  \begin{tabular}[c]{@{}c@{}}Car\\ (6424)\end{tabular} &
  \begin{tabular}[c]{@{}c@{}}Pedestrian\\ (6088)\end{tabular} &
  \begin{tabular}[c]{@{}c@{}}Van\\ (1248)\end{tabular} &
  \begin{tabular}[c]{@{}c@{}}Cyclist\\ (308)\end{tabular} &
  \multicolumn{1}{c}{\begin{tabular}[c]{@{}c@{}}Mean\\ (14068)\end{tabular}} \\ \hline
SC3D & 41.3 / 57.9 & 18.2 / 37.8 & 40.4 / 47.0 & 41.5 / 70.4 & 31.2 / 48.5 \\
P2B  & 56.2 / 72.8 & 28.7 / 49.6 & 40.8 / 48.4 & 32.1 / 44.7 & 42.4 / 60.0 \\
BAT & 60.5 / 77.7 & 42.1 / 70.1 & 52.4 / 67.0 & 33.7 / 45.4 & 51.2 / 72.8 \\
V2B  & 70.5 / 81.3 & 48.3 / 73.5 & 50.1 / 58.0 & 40.8 / 49.7 & 58.4 / 75.2 \\
PTTR & 65.2 / 77.4 & 50.9 / 81.6 & 52.5 / 61.8 & 65.1 / 90.5 & 57.9 / 78.1 \\
STNet & 72.1 / 84.0 & 49.9 / 77.2 & 58.0 / 70.6   & 73.5 / 93.7 & 61.3 / 80.1 \\
M2-Track & 65.5 / 80.8 & 61.5 / 88.2 & 53.8 / 70.7 & 73.2 / 93.5 & 62.9 / 83.4 \\
CXTrack & 69.1 / 81.6 & 67.0 / 91.5 & 60.0 / 71.8 & 74.2 / 94.3 & 67.5 / 85.3 \\
SyncTrack & 73.3 / \underline{85.0} & 54.7 / 80.5 & 60.3 / 70.0 & 73.1 / 93.8 & 64.1 / 81.9 \\
MBPTrack & \underline{73.4} / 84.8 & \underline{68.6} / \underline{93.9} & \underline{61.3} / \underline{72.7} & \underline{76.7} / \underline{94.3} & \underline{70.3} / \underline{87.9} \\ \hline
BEVTrack & \textbf{74.9} / \textbf{86.5} & \textbf{69.5} / \textbf{94.3} & \textbf{66.0} / \textbf{77.2} & \textbf{77.0} / \textbf{94.7} & \textbf{71.8} / \textbf{89.2}\\
Improvement &
  {\color[RGB]{50,203,0} ↑1.5 \color[RGB]{0,0,0} / \color[RGB]{50,203,0} ↑1.5} &
  {\color[RGB]{50,203,0} ↑0.9 \color[RGB]{0,0,0} / \color[RGB]{50,203,0} ↑0.4} &
  {\color[RGB]{50,203,0} ↑4.7 \color[RGB]{0,0,0} / \color[RGB]{50,203,0} ↑4.5} &
  {\color[RGB]{50,203,0} ↑0.3 \color[RGB]{0,0,0} / \color[RGB]{50,203,0} ↑0.4} &
  {\color[RGB]{50,203,0} ↑1.5 \color[RGB]{0,0,0} / \color[RGB]{50,203,0} ↑1.3} \\ 
\hline
\end{tabular}
}
\caption{Comparisons with state-of-the-art methods on KITTI dataset. Success/Precision are used for evaluation. \textbf{Bold} and \underline{underline} denote the best and the second-best scores, respectively.}
\label{Tab:kitti}
\end{table}

\begin{figure*}[!t]
     \includegraphics[width=\textwidth]{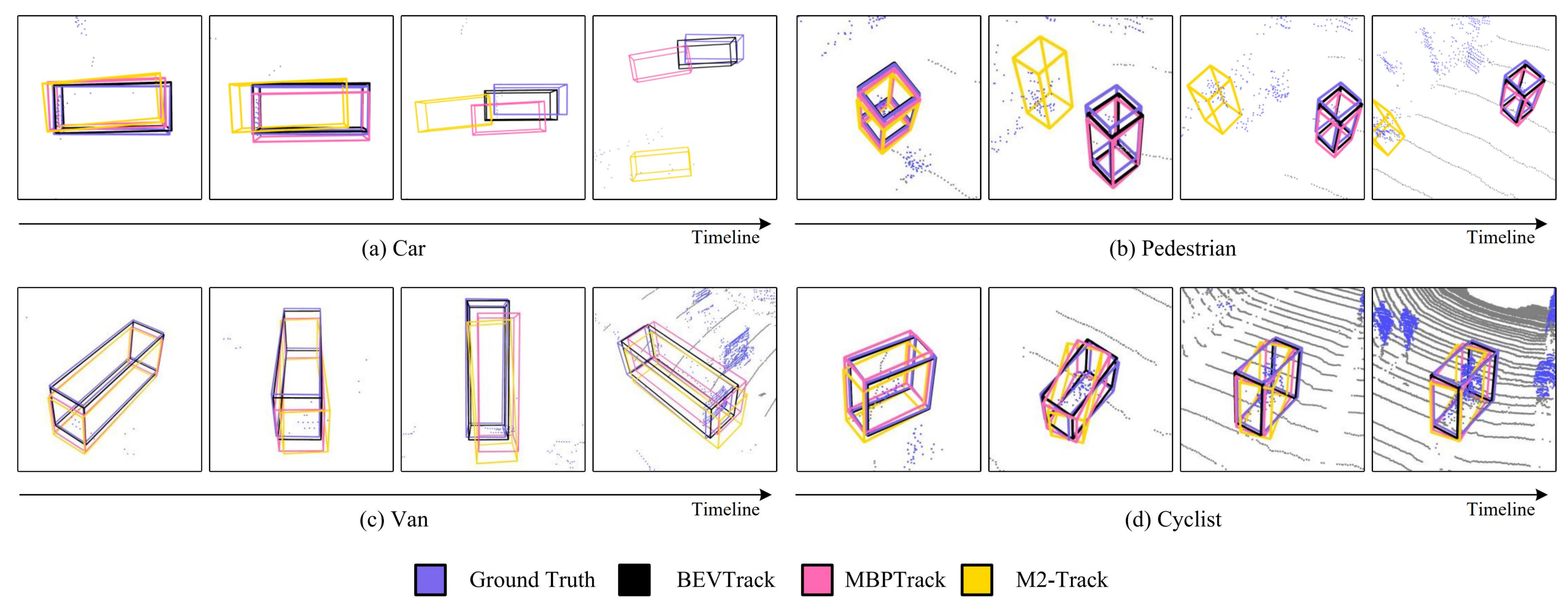}
     \centering
     \caption{Visualization results on different KITTI categories: (a) Car; (b) Pedestrian; (c) Van; (d) Cyclist.}
     \label{fig:vis}
\end{figure*}

\begin{table*}[!h]
\captionsetup{justification=centering}
\centering
\resizebox{\textwidth}{!}{
\renewcommand{\arraystretch}{1.1}
\begin{tabular}{c|c|c|c|c|c|c|c}
\hline
Method &
  \begin{tabular}[c]{@{}c@{}}Car\\ (64,159)\end{tabular} &
  \begin{tabular}[c]{@{}c@{}}Pedestrian\\ (33,227)\end{tabular} &
  \begin{tabular}[c]{@{}c@{}}Truck\\ (13,587)\end{tabular} &
  \begin{tabular}[c]{@{}c@{}}Trailer\\ (3,352)\end{tabular} &
  \begin{tabular}[c]{@{}c@{}}Bus\\ (2,953)\end{tabular} &
  \begin{tabular}[c]{@{}c@{}}Mean\\ by Class\end{tabular} &
  \begin{tabular}[c]{@{}c@{}}Mean\\ (117,278)\end{tabular} \\ \hline
SC3D     & 22.31 / 21.93 & 11.29 / 12.65 & 35.28 / 28.12 & 35.28 / 28.12 & 29.35 / 24.08 & 25.78 / 22.90 & 20.70 / 20.20 \\
P2B      & 38.81 / 43.18 & 28.39 / 52.24 & 48.96 / 40.05 & 48.96 / 40.05 & 32.95 / 27.41 & 38.41 / 40.90 & 36.48 / 45.08 \\
PTT      & 41.22 / 45.26 & 19.33 / 32.03 & 50.23 / 48.56 & 51.70 / 46.50 & 39.40 / 36.70 & 40.38 / 41.81 & 36.33 / 41.72 \\
BAT      & 40.73 / 43.29 & 28.83 / 53.32 & 52.59 / 44.89 & 52.59 / 44.89 & 35.44 / 28.01 & 40.59 / 42.42 & 38.10 / 45.71 \\
PTTR     & 51.89 / 58.61 & 29.90 / 45.09 & 45.30 / 44.74 & 45.87 / 38.36 & 43.14 / 37.74 & 43.22 / 44.91 & 44.50 / 52.07 \\
M2-Track  & 55.85 / 65.09& 32.10 / 60.92 &57.36 / 59.54 & 57.61 / 58.26 & 51.39 / 51.44 & 50.86 / 59.05 & 49.23 / 62.73 \\
MBPTrack & \underline{62.47} / \underline{70.41}  & \underline{45.32} / \underline{74.03} & \underline{62.18} / \underline{63.31} & \underline{65.14} / \underline{61.33} & \underline{55.41} / \underline{51.76} & \underline{58.10} / \underline{64.17}  & \underline{57.48} / \underline{69.88} \\ \hline
BEVTrack & \textbf{64.31 / 71.14}& \textbf{46.28 / 76.77}& \textbf{66.83 / 67.04}& \textbf{74.54 / 71.62}& \textbf{61.09 / 56.68}& 
\textbf{62.61 / 68.65}&\textbf{59.71 / 71.19}\\
Improvement &
  {\color[RGB]{50,203,0} ↑1.84 \color[RGB]{0,0,0} / \color[RGB]{50,203,0} ↑0.73} &
  {\color[RGB]{50,203,0} ↑0.96 \color[RGB]{0,0,0} / \color[RGB]{50,203,0} ↑2.74} &
  {\color[RGB]{50,203,0} ↑4.65 \color[RGB]{0,0,0} / \color[RGB]{50,203,0} ↑3.73} &
  {\color[RGB]{50,203,0} ↑9.40 \color[RGB]{0,0,0} / \color[RGB]{50,203,0} ↑10.29} &
  {\color[RGB]{50,203,0} ↑5.68 \color[RGB]{0,0,0} / \color[RGB]{50,203,0} ↑4.92} &
  {\color[RGB]{50,203,0} ↑4.51 \color[RGB]{0,0,0} / \color[RGB]{50,203,0} ↑4.48} &
  {\color[RGB]{50,203,0} ↑2.23 \color[RGB]{0,0,0} / \color[RGB]{50,203,0} ↑1.31} \\ 
\hline
\end{tabular}
}
\caption{Comparisons with the state-of-the-art methods on NuScenes dataset.} 
\label{Tab:NuScenes}
\end{table*}

\begin{table}[!t]
\centering
\resizebox{\linewidth}{!}{
\renewcommand{\arraystretch}{1.1}
\begin{tabular}{c|ccc}
\hline
Method   & \begin{tabular}[c]{@{}c@{}}Vehicle\\ 1,057,651\end{tabular} & \begin{tabular}[c]{@{}c@{}}Pedestrian\\ 510,533\end{tabular} & \begin{tabular}[c]{@{}c@{}}Mean\\ 1,568,184\end{tabular} \\ \hline
P2B      & 28.32 / 35.41                                               & 15.60 / 29.56                                                & 24.18 / 33.51                                            \\
BAT      & 35.62 / 44.15                                               & 22.05 / 36.79                                                & 31.20 / 41.75                                            \\
M2-Track & \underline{43.62} / \underline{61.64}                                               & \underline{42.10} / \underline{67.31}                                                & \underline{43.13} / \underline{63.48}                                            \\ \hline
BEVTrack & \textbf{70.05} / \textbf{80.05}                                               & \textbf{45.93} / \textbf{72.41}                                                & \textbf{62.20} / \textbf{77.56}                                            \\ 

Improvement &
  {\color[RGB]{50,203,0} ↑26.43 \color[RGB]{0,0,0} / \color[RGB]{50,203,0} ↑18.41} &
  {\color[RGB]{50,203,0} ↑3.83 \color[RGB]{0,0,0} / \color[RGB]{50,203,0} ↑5.10} &
  {\color[RGB]{50,203,0} ↑19.07 \color[RGB]{0,0,0} / \color[RGB]{50,203,0} ↑14.08} \\ 
\hline
\end{tabular}
}
\caption{Comparisons with the state-of-the-art methods on Waymo Open Dataset.}
\label{Tab:waymo_full}
\end{table}

\begin{figure}[!t]
    \includegraphics[width=\linewidth]{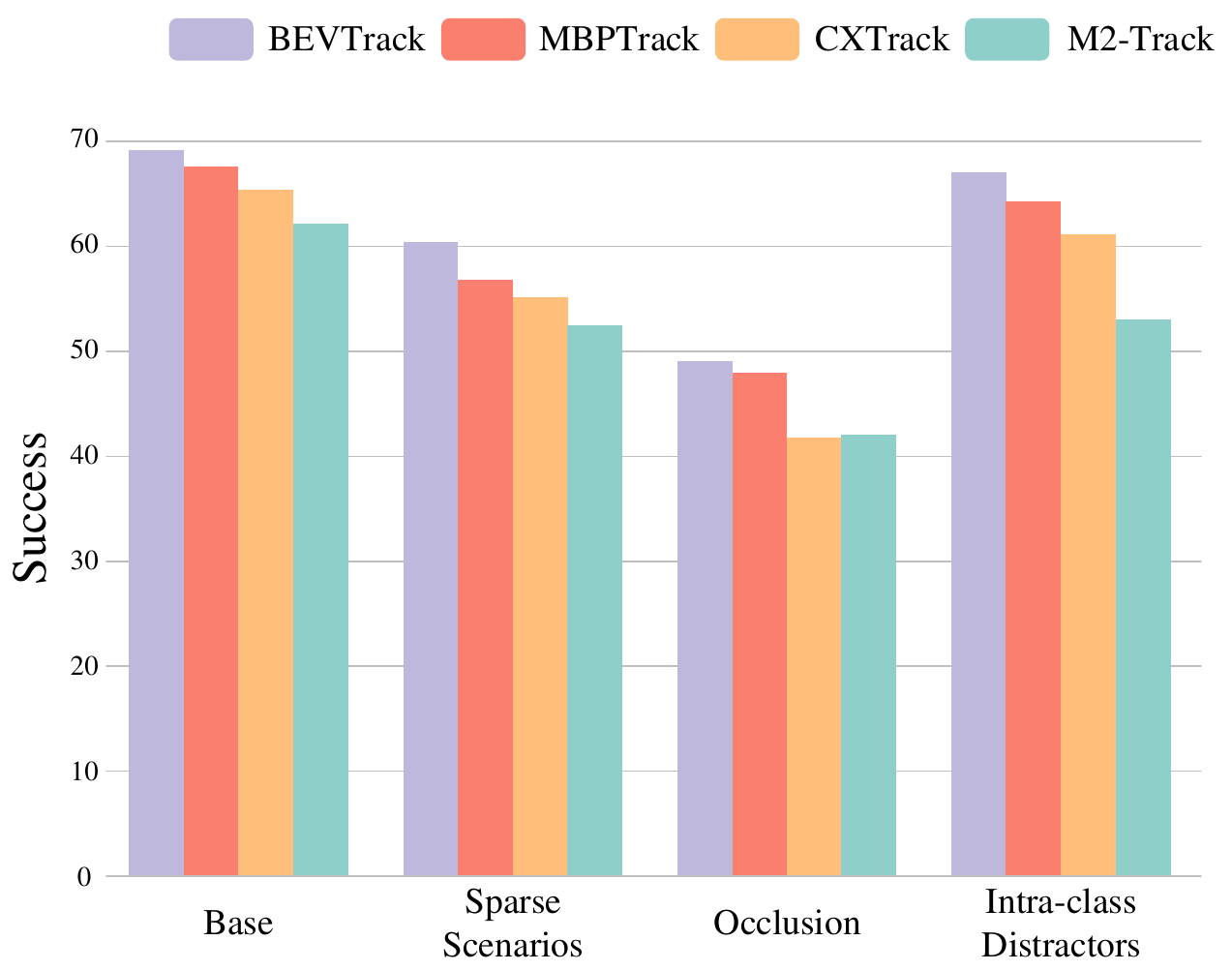}
    \caption{Robustness under different challenging scenes on KITTI Pedestrian category.}
    \label{fig:robustness}
\end{figure}

\subsection{Comparison with State-of-the-art Methods}
\paragraph{Results on KITTI.} We present a comprehensive comparison of BEVTrack with the previous state-of-the-art approaches on KITTI, namely SC3D~\cite{sc3d}, P2B~\cite{P2B}, BAT~\cite{BAT}, V2B~\cite{V2B}, PTTR~\cite{PTTR}, STNet~\cite{STNet}, M2-Track~\cite{M2Track}, CXTrack~\cite{CXTrack}, SyncTrack~\cite{synctrack}, and MBPTrack~\cite{mbptrack}. As shown in Tab.~\ref{Tab:kitti}, BEVTrack surpasses current tracking methods with a significant improvement across all categories on KITTI in \textit{Success} and \textit{Precision} metrics. 
We also visualize the tracking results on four KITTI categories for qualitative comparisons. As shown in Fig.~\ref{fig:vis}, BEVTrack generates more robust and accurate tracking predictions.

For a more comprehensive evaluation of effectiveness in challenging scenarios, we present results from various methods in cases characterized by sparse scenarios, occlusion, and intra-class distractors. To verify the effectiveness in sparse scenarios, we follow V2B~\cite{V2B} to select sparse scenes for evaluation according to the number of points lying in the target bounding boxes in the test set. For analysis of the impact of intra-class distractors, we pick out scenes that contain Pedestrian distractors close to the target from the KITTI test split. For the occlusion situations, we filter out scenes of occlusion scores less than one in the KITTI test set. 

As shown in Fig.~\ref{fig:robustness}, each column from left to right represents the cases of base setting, sparse scenarios, occlusion, and intra-class distractors respectively on the Pedestrian category of the KITTI dataset. Our proposed BEVTrack is more robust to these challenging scenarios than other methods.

\paragraph{Results on Nuscenes.} NuScenes poses a more formidable challenge for the 3D SOT task compared to KITTI, primarily attributable to its larger data volumes and sparser annotations (2Hz for NuScenes versus 10Hz for KITTI and Waymo Open Dataset). Subsequently, following the methodology established by M2-Track, we perform a comparative evaluation on the NuScenes dataset against prior methodologies, namely SC3D~\cite{sc3d}, P2B~\cite{P2B}, PTT~\cite{PTT}, BAT~\cite{BAT}, PTTR~\cite{PTTR}, M2-Track~\cite{M2Track}, and MBPTrack~\cite{mbptrack}. As shown in Tab.~\ref{Tab:NuScenes}, our method achieves a consistent and large performance gain compared with the previous state-of-the-art method, MBPTrack. BEVTrack exhibits superior performance over methods reliant on appearance matching or segmentation, especially in datasets like NuScenes with sparser point clouds. 

\paragraph{Results on WOD.} WOD dataset presents greater challenges compared to KITTI and NuScenes due to its larger data volumes and complexities. Following M2-Track, we conduct training and testing on the respective training and validation sets. We consider trackers that report results under this setting as competitors, including P2B~\cite{P2B}, BAT~\cite{BAT}, and M2-Track~\cite{M2Track}. As illustrated in Tab.~\ref{Tab:waymo_full}, BEVTrack consistently outperforms all competitors on average, particularly excelling in the \textit{Vehicle} category. Remarkably, BEVTrack, with its simple design, demonstrates even greater potential on this large dataset.

\begin{table}[tp]
\centering
\resizebox{0.75\linewidth}{!}{
\renewcommand{\arraystretch}{1.1}
\begin{tabular}{c|ccc}
\hline
Method   & Params & FLOPs & Infer time \\ \hline
CXTrack  & 18.3M  & 4.63G & 16.4ms     \\
MBPTrack & 7.4M   & 2.88G & 11.7ms     \\
BEVTrack & 13.8M  & 0.76G & 4.98ms     \\ \hline
\end{tabular}
}
\caption{Inference efficiency analysis.}
\label{Tab:efficiency}
\end{table}

\paragraph{Inference Speed.} We analyze the efficiency of BEVTrack in Tab.~\ref{Tab:efficiency}. It can be observed that BEVTrack is lightweight with only 0.76G FLOPs and 13.8M parameters. The simple architecture ensures real-time inference of BEVTrack at an impressive speed of 200 FPS on a single NVIDIA GTX 4090 GPU, which is 2.34 times faster than the previously leading method, MBPTrack~\cite{mbptrack}. The simplicity of BEVTrack's pipeline facilitates flexible adjustments in model size by incorporating advanced backbones or devising more effective BMM modules to further enhance performance. 

\subsection{Ablation Studies}

In this section, we analyze and compare each basic design in BEVTrack with other choices used in previous works. For better clarification, we ablate the effects of every design choice by replacing it from the proposed BEVTrack.

\begin{table}[t]
\centering
\resizebox{\linewidth}{!}{
\begin{tabular}{c|cc|cc}
\toprule
 & \multicolumn{2}{c|}{KITTI} & \multicolumn{2}{c}{NuScenes} \\
\cmidrule(lr){2-3} \cmidrule(lr){4-5}
\multirow{-2}{*}{Method} & Car & Ped & Car & Ped \\
\midrule
$Temp$ & 65.7 / 79.3 & 60.8 / 88.5 & 55.3 / 60.2 & 33.6 / 60.5 \\
\midrule
$Seg$ & 71.8 / 84.2 & 65.5 / 92.1 & 58.5 / 65.1 & 40.5 / 65.8 \\
\midrule
$Cx$ & \textbf{74.9} / \textbf{86.5} & \textbf{69.5} / \textbf{94.3} & \textbf{64.3} / \textbf{71.1} & \textbf{46.3} / \textbf{76.8} \\
\bottomrule
\end{tabular}
}
\caption{Ablation study of pre-processing methods.}
\label{Tab:preprocess}
\end{table}

\paragraph{Pre-processing of Input.} 

The input consists of LiDAR-scanned point clouds, with a cropped region where the object may appear. Most existing matching-based methods~\cite{P2B,BAT,3dSiamRPN} use the cropped target template from the previous frame and the full search area in the current frame, referred to as ``$Temp$" in Tab.~\ref{Tab:preprocess}. M2-Track~\cite{M2Track} introduces a motion-centric approach, extracting the target points from their surroundings to feed the motion model, denoted as ``$Seg$". CXTrack~\cite{CXTrack} highlights the loss of contextual information in the previous methods and instead processes two consecutive frames, transforming them into the canonical previous box coordinates, with the region expanded by a margin relative to the object size, referred to as ``$Cx$".

In Tab.~\ref{Tab:preprocess}, we compare the above three pre-processing methods on KITTI and NuScenes. The motion-based tracker BEVTrack performs poorly with the template and search area input (``$Temp$"). While ``$Seg$" yields satisfactory results, it still lags behind "$Cx$". The segmentation network struggles to fully distinguish foreground from background, causing cumulative errors in the motion model. In contrast, ``$Cx$" retains valuable contextual information, leading to the best performance, especially on NuScenes, which uses low-beam LiDAR. As a result, BEVTrack adopts "$Cx$" as its default pre-processing method.

\begin{table}[t]
\centering
\resizebox{\linewidth}{!}{
\begin{tabular}{c|cc|cc}
\toprule
 & \multicolumn{2}{c|}{KITTI} & \multicolumn{2}{c}{NuScenes} \\
\cmidrule(lr){2-3} \cmidrule(lr){4-5}
\multirow{-2}{*}{Repr} & Car & Ped & Car & Ped \\
\midrule
$Point$ & 70.5 / 82.2 & 66.3 / 90.5 & 60.7 / 66.2 & 41.3 / 68.6 \\
\midrule
$Voxel$ & \textbf{74.9} / \textbf{86.5} & \textbf{69.5} / \textbf{94.3} & \textbf{64.3} / \textbf{71.1} & \textbf{46.3} / \textbf{76.8} \\
\bottomrule
\end{tabular}
}
\caption{Ablation study of representation choices.}
\label{Tab:representation}
\end{table}

\begin{figure}[tp]
    \centering
    \includegraphics[width=\linewidth]{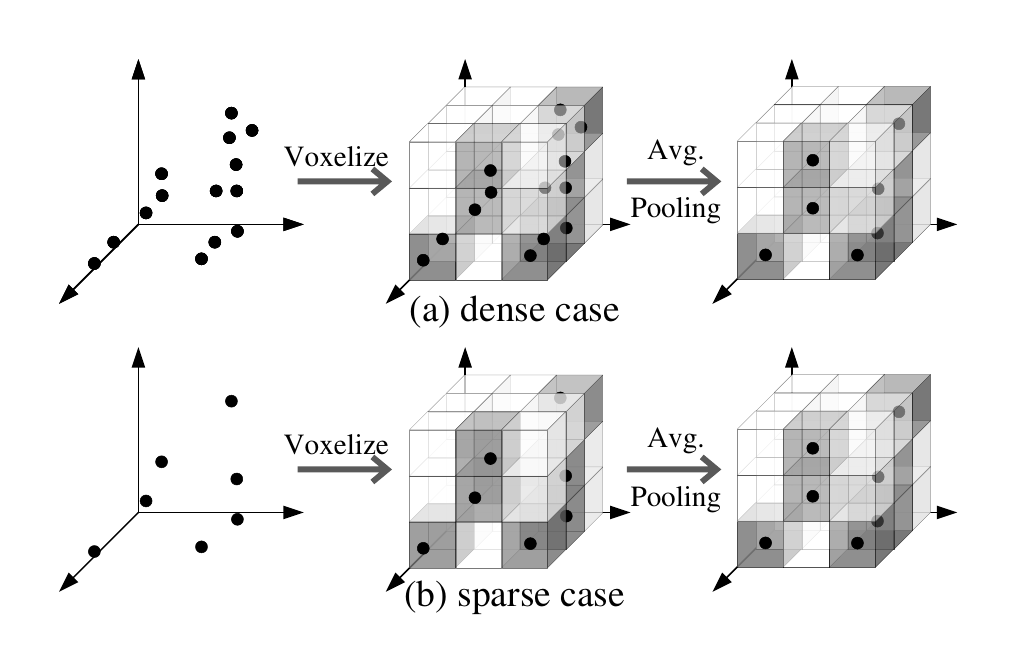}
    \caption{Diagram of voxelization. The density differences of point clouds are alleviated after voxelization.} 
    \label{fig:voxelization}
\end{figure}

\paragraph{Feature representation.} 

We aim to learn a discriminative feature representation from point clouds. While existing 3D SOT methods often rely on point-based representation networks~\cite{PointNet,PointNet++,DGCNN}, LiDAR-scanned point clouds exhibit variable sparsity—dense clustering near objects and sparser distribution at a distance. Effectively learning point features that handle both dense and sparse regions is a significant challenge. To address this, we propose using voxel-based representations~\cite{VoxelNet} instead of point-based ones. As shown in Fig.~\ref{fig:voxelization}, voxel features are less sensitive to point count variations, helping mitigate the sparsity issue.

In Tab.~\ref{Tab:representation}, we compare the ``Point" and ``Voxel" representations on the KITTI and NuScenes datasets. The ``Voxel" setup follows the same approach as BEVTrack. To ensure a fair comparison, the ``Point" method adopts a similar pipeline, including feature extraction, motion modeling, and regression. However, ``Point" uses PointNet++~\cite{PointNet++} for feature extraction, capturing point-level features. To distinguish points across frames, ``Point" adds timestamp encoding to the features, which are then processed by a second PointNet++ for motion modeling. As shown in Tab.~\ref{Tab:representation}, the significant performance gap between ``Point" and ``Voxel" supports our claim above, highlighting the effectiveness of using voxel-based representations in BEVTrack.

\begin{table}[t]
\centering
\resizebox{\linewidth}{!}{
\begin{tabular}{c|cc|cc}
\toprule
 & \multicolumn{2}{c|}{KITTI} & \multicolumn{2}{c}{NuScenes} \\
\cmidrule(lr){2-3} \cmidrule(lr){4-5}
\multirow{-2}{*}{Method} & Car & Ped & Car & Ped \\
\midrule
PointNet & 68.8 / 81.7 & 62.9 / 89.1 & 62.1 / 68.5 & 39.7 / 65.6 \\
\midrule
BMM (ours) & \textbf{74.9} / \textbf{86.5} & \textbf{69.5} / \textbf{94.3} & \textbf{64.3} / \textbf{71.1} & \textbf{46.3} / \textbf{76.8} \\
\bottomrule
\end{tabular}
}
\caption{Ablation study of motion modeling methods.}
\label{Tab:motion}
\end{table}

\paragraph{Motion Modeling.} 

aims to infer the relative target motion (RTM) from consecutive point cloud frames. M2-Track~\cite{M2Track}, the representative motion-based tracker preceding BEVTrack, uses PointNet~\cite{PointNet} to predict RTM. In contrast, we introduce BMM for RTM modeling, as described in Sec.~\ref{subsec:bmm}. BMM offers two key advantages over PointNet: 1) BEV better captures motion features, particularly for horizontal movement typical in autonomous driving; 2) BMM leverages cascaded convolutions to model local features and capture diverse motion patterns.

In Tab.~\ref{Tab:motion}, we compare the two motion modeling methods on KITTI and NuScenes. For the ``PointNet" setting, we restore voxel features to point features using bilinear interpolation and replace the BMM module in BEVTrack with PointNet. As shown, BMM outperforms PointNet by a clear margin, highlighting the importance of local feature learning and BEV representation for motion modeling. Additionally, we observe that the methods differ in sensitivity to background points. According to Tab.~\ref{Tab:preprocess}, BEVTrack performs better without a segmentation network, while M2-Track benefits significantly from it. This suggests that BMM's local feature learning is more robust to background interference.

\begin{table}[t]
\centering
\resizebox{\linewidth}{!}{
\begin{tabular}{c|cc|cc}
\toprule
 & \multicolumn{2}{c|}{KITTI} & \multicolumn{2}{c}{NuScenes} \\
\cmidrule(lr){2-3} \cmidrule(lr){4-5}
\multirow{-2}{*}{Paradigm} & Car & Ped & Car & Ped \\
\midrule
matching-based & 71.5 / 83.4 & 61.8 / 87.4 & 60.0 / 66.8 & 32.7 / 58.8 \\
\midrule
motion-based & \textbf{74.9} / \textbf{86.5} & \textbf{69.5} / \textbf{94.3} & \textbf{64.3} / \textbf{71.1} & \textbf{46.3} / \textbf{76.8} \\
\bottomrule
\end{tabular}
}
\caption{Ablation study of paradigm choices.}
\label{Tab:paradigm}
\end{table}

\paragraph{Paradigm Choices.} 

Most existing 3D SOT methods follow either a matching-based or motion-based paradigm. Matching-based trackers~\cite{P2B,BAT,PTTR} locate targets by comparing the search area with template point clouds. However, LiDAR point clouds are often textureless and incomplete, hindering effective appearance matching. In contrast, motion-based tracking benefits from the absolute scale of target movement in point clouds, simplifying the tracking process. In this paper, BEVTrack adopts the motion-based paradigm.

We investigate the effect of paradigm selection on BEVTrack's performance by designing a matching-based variant that uses the same basic modules for fairness. Details of the variant's architecture can be found in the appendix. As shown in Tab.~\ref{Tab:paradigm}, the ``motion-based" BEVTrack outperforms the matching-based variant by a significant margin. For the pedestrian class, which has similar intra-class distractors often appearing at closer distances, the performance of the matching-based version drops sharply by 7.7/6.9 in success/precision on KITTI and 13.6/18.0 on NuScenes.

\begin{table}[t]
\centering
\resizebox{\linewidth}{!}{
\renewcommand{\arraystretch}{1.1}
\begin{tabular}{c|c|c|c|c|c}
\hline
                   & Car         & Pedestrian  & Van         & Cyclist     & Mean        \\ \hline
G & 68.0 / 80.8& 59.8 / 88.6& 61.8 / 71.9& 69.8 / 92.4& 63.9 / 83.6\\
L  & 69.2 / 81.8& 64.2 / 90.7& 58.9 / 72.2& 73.2 / 93.5& 66.2 / 85.1\\ 
D & \textbf{74.9 / 86.5}& \textbf{69.5 / 94.3}& \textbf{66.0 / 77.2}& \textbf{77.0 / 94.7}& \textbf{71.8 / 89.2}\\ \hline
\end{tabular}
}
\caption{Ablation study of the regression strategy. ``G" refers to regress with Gaussian assumption, ``L" refers to regress with Laplacian assumption, and ``D" refers to regress with distribution-aware tracking strategy.}
\label{Tab:distribution}
\end{table}

\paragraph{Distribution-aware Regression.} 

To investigate the impact of output distribution assumptions on tracking performance, we compare results using different density functions on the KITTI dataset. The Laplacian and Gaussian distributions reduce to standard $l_{1}$ and $l_{2}$ loss when assumed to have constant variances. As shown in Tab.~\ref{Tab:distribution}, the proposed distribution-aware regression approach outperforms regression with $l_{1}$ or $l_{2}$ loss, highlighting the importance of modeling the actual distribution of target motion.

\section{Conclution}
This paper introduces BEVTrack, a simple yet strong baseline for 3D single object tracking (SOT). BEVTrack performs tracking within the Bird's-Eye View representation, thereby effectively exploiting spatial information and capturing motion cues. Additionally, we propose a distribution-aware regression strategy that learns the actual distribution adapted to targets possessing diverse attributes, providing accurate guidance for tracking. Comprehensive experiments conducted on widely recognized benchmarks underscore BEVTrack's efficacy, establishing its superiority over state-of-the-art tracking methods. Furthermore, it achieves a high inference speed of about 200 FPS. We hope this study could provide valuable insights to the tracking community and inspire further research.

\paragraph{Limitation.}

BEVTrack, as a simple baseline, uses only two successive frames, but the sequence of historical trajectories and point clouds holds rich motion data. Modeling object motion from multiple frames could enhance tracking performance. We plan to explore the use of temporal information in BEVTrack for long-term video tracking in future work.

\clearpage

\section*{Acknowledgements}
This work was supported by the National Natural Science Foundation of China (62376080), the Fundamental Research Funds for the Provincial Universities of Zhejiang (GK259909299001-005), the Basic Research Projects of China (JCKY2022211C004, JCKY2023415C009), and the Zhejiang Key Laboratory of Optoelectronic Intelligent Imaging and Aerospace Sensing. Thanks Jiahao Nie and Jinlong Fan for their helpful discussions.

\bibliographystyle{named}
\bibliography{ijcai25}

\begin{thebibliography}{}

\bibitem[\protect\citeauthoryear{Caesar \bgroup \em et al.\egroup }{2020}]{NuScenes}
Holger Caesar, Varun Bankiti, Alex~H Lang, Sourabh Vora, Venice~Erin Liong, Qiang Xu, Anush Krishnan, Yu~Pan, Giancarlo Baldan, and Oscar Beijbom.
\newblock Nuscenes: A multimodal dataset for autonomous driving.
\newblock In {\em Proceedings of the IEEE/CVF Conference on Computer Vision and Pattern Recognition}, pages 11621--11631, 2020.

\bibitem[\protect\citeauthoryear{Chen \bgroup \em et al.\egroup }{2023}]{crossd}
Guanlin Chen, Pengfei Zhu, Bing Cao, Xing Wang, and Qinghua Hu.
\newblock Cross-drone transformer network for robust single object tracking.
\newblock {\em IEEE Transactions on Circuits and Systems for Video Technology}, 33(9):4552--4563, 2023.

\bibitem[\protect\citeauthoryear{Dinh \bgroup \em et al.\egroup }{2016}]{NVP}
Laurent Dinh, Jascha Sohl-Dickstein, and Samy Bengio.
\newblock Density estimation using real nvp.
\newblock In {\em Proceedings of the International Conference on Learning Representations}, 2016.

\bibitem[\protect\citeauthoryear{Duffhauss and Baur}{2020}]{pillarflownet}
Fabian Duffhauss and Stefan~A Baur.
\newblock Pillarflownet: A real-time deep multitask network for lidar-based 3d object detection and scene flow estimation.
\newblock In {\em 2020 IEEE/RSJ International Conference on Intelligent Robots and Systems}, pages 10734--10741. IEEE, 2020.

\bibitem[\protect\citeauthoryear{Fang \bgroup \em et al.\egroup }{2021}]{3dSiamRPN}
Zheng Fang, Sifan Zhou, Yubo Cui, and Sebastian Scherer.
\newblock 3d-siamrpn: An end-to-end learning method for real-time 3d single object tracking using raw point cloud.
\newblock {\em IEEE Sensors Journal}, page 4995–5011, 2021.

\bibitem[\protect\citeauthoryear{Geiger \bgroup \em et al.\egroup }{2012}]{KITTI}
Andreas Geiger, Philip Lenz, and Raquel Urtasun.
\newblock Are we ready for autonomous driving? the kitti vision benchmark suite.
\newblock In {\em Proceedings of the IEEE Conference on Computer Vision and Pattern Recognition}, pages 3354--3361, 2012.

\bibitem[\protect\citeauthoryear{Giancola \bgroup \em et al.\egroup }{2019}]{sc3d}
Silvio Giancola, Jesus Zarzar, and Bernard Ghanem.
\newblock Leveraging shape completion for 3d siamese tracking.
\newblock In {\em Proceedings of the IEEE/CVF Conference on Computer Vision and Pattern Recognition}, pages 1359--1368, 2019.

\bibitem[\protect\citeauthoryear{Graham and Van~der Maaten}{2017}]{submanifold}
Benjamin Graham and Laurens Van~der Maaten.
\newblock Submanifold sparse convolutional networks.
\newblock {\em arXiv preprint arXiv:1706.01307}, 2017.

\bibitem[\protect\citeauthoryear{Graham \bgroup \em et al.\egroup }{2018}]{sparse_conv}
Benjamin Graham, Martin Engelcke, and Laurens Van Der~Maaten.
\newblock 3d semantic segmentation with submanifold sparse convolutional networks.
\newblock In {\em Proceedings of the IEEE Conference on Computer Vision and Pattern Recognition}, pages 9224--9232, 2018.

\bibitem[\protect\citeauthoryear{Hui \bgroup \em et al.\egroup }{2021}]{V2B}
Le~Hui, Lingpeng Wang, Mingmei Cheng, Jin Xie, and Jian Yang.
\newblock 3d siamese voxel-to-bev tracker for sparse point clouds.
\newblock {\em Advances in Neural Information Processing Systems}, 34:28714--28727, 2021.

\bibitem[\protect\citeauthoryear{Hui \bgroup \em et al.\egroup }{2022}]{STNet}
Le~Hui, Lingpeng Wang, Linghua Tang, Kaihao Lan, Jin Xie, and Jian Yang.
\newblock 3d siamese transformer network for single object tracking on point clouds.
\newblock In {\em Proceedings of the European Conference on Computer Vision}, pages 293--310, 2022.

\bibitem[\protect\citeauthoryear{Kristan \bgroup \em et al.\egroup }{2016}]{metrics}
Matej Kristan, Jiri Matas, Ales Leonardis, Tomas Vojir, Roman Pflugfelder, Gustavo Fernandez, Georg Nebehay, Fatih Porikli, and Luka Cehovin.
\newblock A novel performance evaluation methodology for single-target trackers.
\newblock {\em IEEE Transactions on Pattern Analysis and Machine Intelligence}, page 2137–2155, 2016.

\bibitem[\protect\citeauthoryear{Li \bgroup \em et al.\egroup }{2021}]{RLE}
Jiefeng Li, Siyuan Bian, Ailing Zeng, Can Wang, Bo~Pang, Wentao Liu, and Cewu Lu.
\newblock Human pose regression with residual log-likelihood estimation.
\newblock In {\em Proceedings of the IEEE/CVF International Conference on Computer Vision}, pages 11025--11034, 2021.

\bibitem[\protect\citeauthoryear{Ma \bgroup \em et al.\egroup }{2023}]{synctrack}
Teli Ma, Mengmeng Wang, Jimin Xiao, Huifeng Wu, and Yong Liu.
\newblock Synchronize feature extracting and matching: A single branch framework for 3d object tracking.
\newblock In {\em Proceedings of the IEEE/CVF International Conference on Computer Vision}, pages 9953--9963, 2023.

\bibitem[\protect\citeauthoryear{Pieropan \bgroup \em et al.\egroup }{2015}]{rgbd2}
Alessandro Pieropan, Niklas Bergstr{\"o}m, Masatoshi Ishikawa, and Hedvig Kjellstr{\"o}m.
\newblock Robust 3d tracking of unknown objects.
\newblock In {\em Proceedings of the IEEE International Conference on Robotics and Automation}, pages 2410--2417, 2015.

\bibitem[\protect\citeauthoryear{Qi \bgroup \em et al.\egroup }{2017a}]{PointNet}
Charles~R Qi, Hao Su, Kaichun Mo, and Leonidas~J Guibas.
\newblock Pointnet: Deep learning on point sets for 3d classification and segmentation.
\newblock In {\em Proceedings of the IEEE Conference on Computer Vision and Pattern Recognition}, pages 652--660, 2017.

\bibitem[\protect\citeauthoryear{Qi \bgroup \em et al.\egroup }{2017b}]{PointNet++}
Charles~Ruizhongtai Qi, Li~Yi, Hao Su, and Leonidas~J Guibas.
\newblock Pointnet++: Deep hierarchical feature learning on point sets in a metric space.
\newblock {\em Advances in Neural Information Processing Systems}, 30, 2017.

\bibitem[\protect\citeauthoryear{Qi \bgroup \em et al.\egroup }{2019}]{VoteNet}
Charles~R Qi, Or~Litany, Kaiming He, and Leonidas~J Guibas.
\newblock Deep hough voting for 3d object detection in point clouds.
\newblock In {\em Proceedings of the IEEE/CVF International Conference on Computer Vision}, pages 9277--9286, 2019.

\bibitem[\protect\citeauthoryear{Qi \bgroup \em et al.\egroup }{2020}]{P2B}
Haozhe Qi, Chen Feng, Zhiguo Cao, Feng Zhao, and Yang Xiao.
\newblock P2b: point-to-box network for 3d object tracking in point clouds.
\newblock In {\em Proceedings of the IEEE/CVF Conference on Computer Vision and Pattern Recognition}, pages 6329--6338, 2020.

\bibitem[\protect\citeauthoryear{Shan \bgroup \em et al.\egroup }{2021}]{PTT}
Jiayao Shan, Sifan Zhou, Zheng Fang, and Yubo Cui.
\newblock Ptt: Point-track-transformer module for 3d single object tracking in point clouds.
\newblock In {\em Proceedings of the IEEE/RSJ International Conference on Intelligent Robots and Systems}, pages 1310--1316, 2021.

\bibitem[\protect\citeauthoryear{Spinello \bgroup \em et al.\egroup }{2010}]{rgbd3}
Luciano Spinello, Kai Arras, Rudolph Triebel, and Roland Siegwart.
\newblock A layered approach to people detection in 3d range data.
\newblock In {\em Proceedings of the AAAI Conference on Artificial Intelligence}, pages 1625--1630, 2010.

\bibitem[\protect\citeauthoryear{Sun \bgroup \em et al.\egroup }{2020}]{waymo}
Pei Sun, Henrik Kretzschmar, Xerxes Dotiwalla, Aurelien Chouard, Vijaysai Patnaik, Paul Tsui, James Guo, Yin Zhou, Yuning Chai, Benjamin Caine, Vijay Vasudevan, Wei Han, Jiquan Ngiam, Hang Zhao, Aleksei Timofeev, Scott Ettinger, Maxim Krivokon, Amy Gao, Aditya Joshi, Yu~Zhang, Jonathon Shlens, Zhifeng Chen, and Dragomir Anguelov.
\newblock Scalability in perception for autonomous driving: waymo open dataset.
\newblock In {\em Proceedings of the IEEE/CVF Conference on Computer Vision and Pattern Recognition}, pages 2446--2454, 2020.

\bibitem[\protect\citeauthoryear{Vaswani \bgroup \em et al.\egroup }{2017}]{transformer}
Ashish Vaswani, Noam Shazeer, Niki Parmar, Jakob Uszkoreit, Llion Jones, AidanN. Gomez, Lukasz Kaiser, and Illia Polosukhin.
\newblock Attention is all you need.
\newblock {\em Advances in Neural Information Processing Systems}, 30, 2017.

\bibitem[\protect\citeauthoryear{Wang \bgroup \em et al.\egroup }{2019}]{DGCNN}
Yue Wang, Yongbin Sun, Ziwei Liu, Sanjay~E. Sarma, Michael~M. Bronstein, and Justin~M. Solomon.
\newblock Dynamic graph cnn for learning on point clouds.
\newblock {\em ACM Transactions on Graphics}, page 1–12, 2019.

\bibitem[\protect\citeauthoryear{Xu \bgroup \em et al.\egroup }{2023a}]{CXTrack}
Tian-Xing Xu, Yuan-Chen Guo, Yu-Kun Lai, and Song-Hai Zhang.
\newblock Cxtrack: Improving 3d point cloud tracking with contextual information.
\newblock In {\em Proceedings of the IEEE/CVF Conference on Computer Vision and Pattern Recognition}, pages 1084--1093, 2023.

\bibitem[\protect\citeauthoryear{Xu \bgroup \em et al.\egroup }{2023b}]{mbptrack}
Tian-Xing Xu, Yuan-Chen Guo, Yu-Kun Lai, and Song-Hai Zhang.
\newblock Mbptrack: Improving 3d point cloud tracking with memory networks and box priors.
\newblock In {\em Proceedings of the IEEE/CVF International Conference on Computer Vision}, pages 9911--9920, 2023.

\bibitem[\protect\citeauthoryear{Yin \bgroup \em et al.\egroup }{2021}]{centerpoint}
Tianwei Yin, Xingyi Zhou, and Philipp Krahenbuhl.
\newblock Center-based 3d object detection and tracking.
\newblock In {\em Proceedings of the IEEE/CVF Conference on Computer Vision and Pattern Recognition}, pages 11784--11793, 2021.

\bibitem[\protect\citeauthoryear{Zheng \bgroup \em et al.\egroup }{2021}]{BAT}
Chaoda Zheng, Xu~Yan, Jiantao Gao, Weibing Zhao, Wei Zhang, Zhen Li, and Shuguang Cui.
\newblock Box-aware feature enhancement for single object tracking on point clouds.
\newblock In {\em Proceedings of the IEEE/CVF International Conference on Computer Vision}, pages 13199--13208, 2021.

\bibitem[\protect\citeauthoryear{Zheng \bgroup \em et al.\egroup }{2022}]{M2Track}
Chaoda Zheng, Xu~Yan, Haiming Zhang, Baoyuan Wang, Shenghui Cheng, Shuguang Cui, and Zhen Li.
\newblock Beyond 3d siamese tracking: A motion-centric paradigm for 3d single object tracking in point clouds.
\newblock In {\em Proceedings of the IEEE/CVF International Conference on Computer Vision}, pages 8111--8120, 2022.

\bibitem[\protect\citeauthoryear{Zhou and Tuzel}{2018}]{VoxelNet}
Yin Zhou and Oncel Tuzel.
\newblock Voxelnet: End-to-end learning for point cloud based 3d object detection.
\newblock In {\em Proceedings of the IEEE/CVF Conference on Computer Vision and Pattern Recognition}, pages 4490--4499, 2018.

\bibitem[\protect\citeauthoryear{Zhou \bgroup \em et al.\egroup }{2022}]{PTTR}
Changqing Zhou, Zhipeng Luo, Yueru Luo, Tianrui Liu, Liang Pan, Zhongang Cai, Haiyu Zhao, and Shijian Lu.
\newblock Pttr: Relational 3d point cloud object tracking with transformer.
\newblock In {\em Proceedings of the IEEE/CVF Conference on Computer Vision and Pattern Recognition}, pages 8531--8540, 2022.

\end{thebibliography}

\end{document}